\documentclass[review]{elsarticle}

\usepackage{hyperref}

\journal{Glottometrics}

\biboptions{authoryear}

\usepackage{changepage}

\usepackage{color}
\usepackage{tabularx}
\usepackage{etoolbox}

\newtoggle{anonymous}
\togglefalse{anonymous}

\newcommand{\LastAccessed}{Last accessed 17 February 2022.}

\begin{document}

\begin{frontmatter}

\title{Memory limitations are hidden in grammar}

\author[Galicia]{Carlos G\'omez-Rodr\'iguez \href{https://orcid.org/0000-0003-0752-8812}{(ORCID 0000-0003-0752-8812)} }

\author[US,Denmark]{Morten H. Christiansen \href{https://orcid.org/0000-0002-3850-0655}{(ORCID 0000-0002-3850-0655)}}
\author[Catalunya]{Ramon Ferrer-i-Cancho \href{https://orcid.org/0000-0002-7820-923X}{(ORCID 0000-0002-7820-923X)}\corref{mycorrespondingauthor}}
\cortext[mycorrespondingauthor]{Corresponding author. 
E-mail: rferrericancho@cs.upc.edu. Phone: +34 934134028. Departament de Ci\`encies de la Computaci\'o, Universitat Polit\`ecnica de Catalunya (UPC), Campus Nord, Edifici Omega, Despatx S124, c/ Jordi Girona Salgado 1-3, 08034 Barcelona, Catalonia, Spain.}

\address[Galicia]{Universidade da Coru\~na, CITIC, FASTPARSE Lab, LyS Research Group, Depto. de Ciencias de la Computaci\'on y Tecnolog\'ias de la Informaci\'on, A Coru\~na, Spain.}
\address[US]{Department of Psychology, Cornell University, Ithaca, NY, USA.}
\address[Denmark]{Interacting Minds Centre and School of Communication and Culture, Nobelparken, Aarhus University, Denmark}
\address[Catalunya]{Complexity and Quantitative Linguistics Lab, LARCA Research Group, Departament de Ci\`encies de la Computaci\'o, Universitat Polit\`ecnica de Catalunya (UPC), Barcelona, Catalonia, Spain.}

\begin{abstract}
The ability to produce and understand an unlimited number of different sentences is a hallmark of human language. Linguists have sought to define the essence of this generative capacity using formal grammars that describe the syntactic dependencies between constituents, independent of the computational limitations of the human brain. Here, we evaluate this independence assumption by sampling sentences uniformly from the space of possible syntactic structures. We find that the average dependency distance between syntactically related words, a proxy for memory limitations, is less than expected by chance in a collection of state-of-the-art classes of dependency grammars. Our findings indicate that memory limitations have permeated grammatical descriptions, suggesting that 
it may be impossible to build a parsimonious theory of human linguistic productivity independent of non-linguistic cognitive constraints.
\end{abstract}


\end{frontmatter}


\section{Introduction}


An often celebrated aspect of human language is its capacity to produce an unbounded number of different sentences \citep{Chomsky1965, Miller1999a}. For many centuries, the goal of linguistics has been to capture this capacity by a formal description---a grammar---consisting of a systematic set of rules and/or principles that determine which sentences are part of a given language and which are not \citep{Bod2013a}. Over the years, these formal grammars have taken many forms but common to them all is the assumption that they capture the idealized linguistic competence of a native speaker/hearer, independent of any memory limitations or other non-linguistic cognitive constraints \citep{Chomsky1965,Miller1999a}. These abstract formal descriptions have come to play a foundational role in the language sciences, from linguistics, psycholinguistics, and neurolinguistics \citep{Hauser2002,Pinker2003a} to computer science, engineering, and machine learning \citep{klein03,dyer-etal-2016-recurrent,GomShiLee2018global}. 
Despite evidence that processing difficulty underpins the unacceptability of certain sentences \citep{Morrill2010a,Hawkins2004a}, the cognitive independence assumption that is a defining feature of linguistic competence has not been examined in a systematic way using the tools of formal grammar. It is therefore unclear whether these supposedly idealized descriptions of language are free of non-linguistic cognitive constraints, such as memory limitations.

If the cognitive independence assumption should turn out not to hold, then it would have wide-spread theoretical and practical implications for our understanding of human linguistic productivity. It would require a reappraisal of key parts of linguistic theory that hitherto have posed formidable challenges for explanations of language processing, acquisition and evolution \citep{Gold1967,Hauser2002,Pinker2003a}---pointing to new ways of thinking about language that may simplify the problem space considerably by making it possible to explain apparently arbitrary aspects of
linguistic structure in terms of general learning and processing biases \citep{Christiansen2008a,Gomez2016a}. In terms of practical ramifications, engineers may benefit from building human cognitive limitations directly into their natural language processing systems, so as to better mimic human language skills and thereby improve performance. Here, we therefore evaluate the cognitive independence assumption using a state-of-the-art grammatical framework, dependency grammar \citep{NivreDGA05}, to search for possible hidden memory constraints in these formal, idealized descriptions of natural language.

\begin{figure}
\centering
\includegraphics[width = \textwidth]{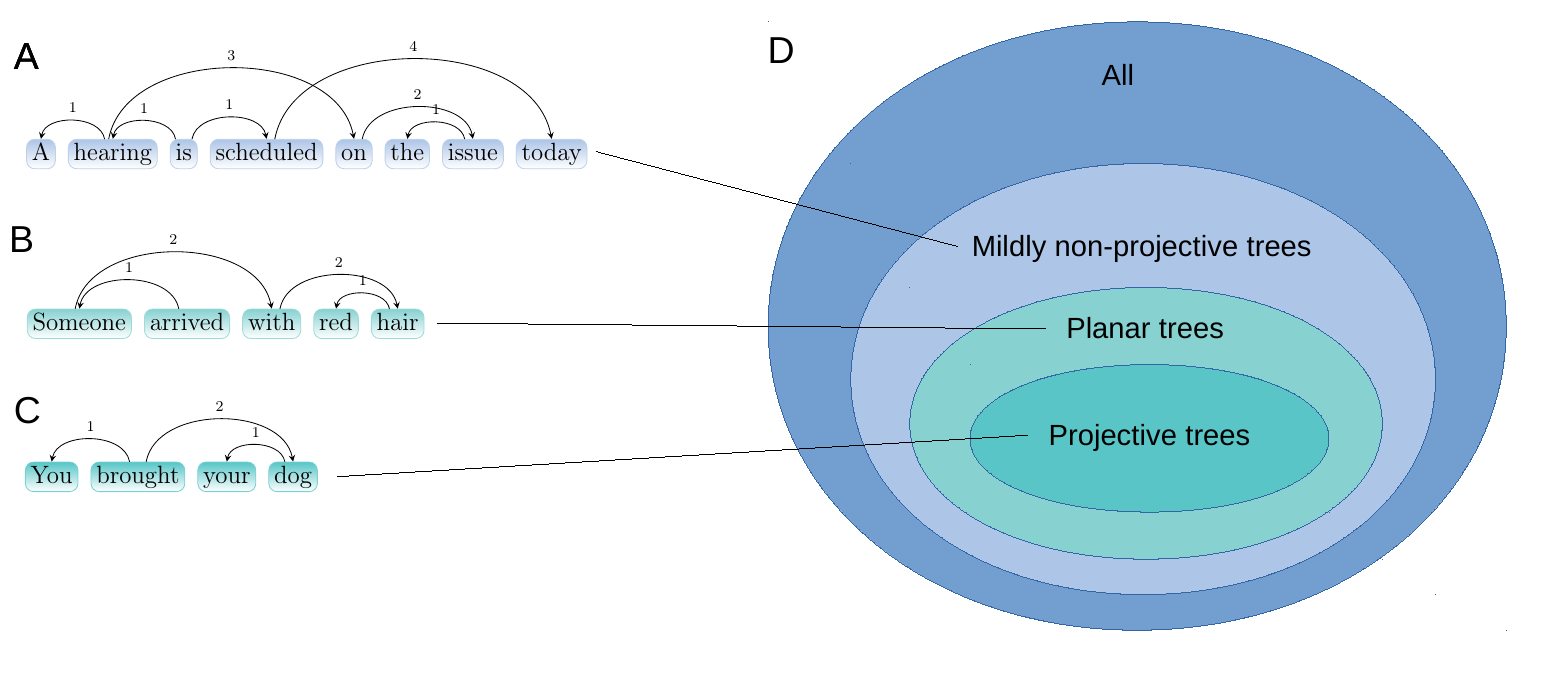}
\caption{\label{syntactic_dependency_trees_figure} Examples of syntactic dependency structures. Arcs indicate syntactic dependencies from a head to its dependent and are labelled with the distance between them (distance is measured in words; consecutive words are at distance 1). $n$ is the number of words of the sentence, $D$ is the sum of dependency distances and $\left<d \right> = D/(n-1)$ is the average dependency distance. 
A. A mildly non-projective tree from the classes $1EC$ and $MH_4$ (adapted from \cite{nivre09acl}) where $n = 8$ and $\left<d \right> = 13/7 \approx 1.85$.
B. A planar but non-projective tree where $n = 5$ and $\left<d \right> = 3/2$ (adapted from \cite{Gross2009a}).
C. A projective tree (adapted from \cite{Gross2009a}) where $n = 4$ and $\left<d \right> = 4/3$.  
D. A diagram of the superset relationships between projective, planar, mildly non-projective and unrestricted (all) syntactic dependency structures.}
\end{figure}

In dependency grammar, the syntactic structure of a sentence is defined by two components. First, a directed graph where vertices are words and arcs indicate syntactic dependencies between a head and its dependent. Such a graph has a root (a vertex that receives no edges) and edges are oriented away from the root (Fig \ref{syntactic_dependency_trees_figure}).  
Second, the linear arrangement of the vertices of the graph (defined by the sequential order of the words in a sentence).
Thus, syntactic dependency structures constitute a particular kind of spatial network where the graph is embedded in one dimension \citep{Barthelemy2018a}, a correspondence that has led to the development of syntactic theory from a network theory standpoint \citep{Gomez2016a}. 

Dependency grammar is an important framework for various reasons. First, categorial grammar defines the syntactic structure of a sentence as dependency grammar \citep{Morrill2010a}. Second,  equivalences exist between certain formalisms of dependency grammar and constituency grammar \citep{Gaifman1965,kahane-mazziotta-2015-syntactic}. 
Third, there has been an evolution of minimalism towards dependency grammar \citep{Osborne2011a}. Finally, dependency grammar has become a {\em de facto} standard in computational linguistics \citep{kubler09book}. 


To delimit the set of possible grammatical descriptions, various classes or sets of syntactic dependency structures have been proposed.
These classes can be seen as filters on the possible linear arrangements of a given tree.
Here, we consider four main classes. 
First, consider planar structures, where edges do not cross when drawn above the words of
the sentence. The structure in Figs \ref{syntactic_dependency_trees_figure}B-C are planar while that of Fig \ref{syntactic_dependency_trees_figure}A is not.  
Second, we have projective structures, the most well-known class. 
A dependency tree is projective if, and only if, it is planar and its
root is not covered by any dependency (Fig \ref{syntactic_dependency_trees_figure}C). 
Third, there are mildly non-projective structures, comprising the union of planar structures and additional structures 
with further (but slight) deviations from projectivity, e.g., by having a low number of edge crossings (Fig \ref{syntactic_dependency_trees_figure}A).
Finally, the class of all structures, that has no constraints on the possible structures. 

Fig. \ref{syntactic_dependency_trees_figure}D shows the inclusion relationships among these classes. However, the whole picture, encompassing state-of-the-art classes is more complex. Mildly non-projective structures are not actually a class but a family of classes. 
We have selected three representative classes: $MH_k$, $WG_1$ and $1EC$ structures, that are supersets of projective structures but whose definition is more complex (see Methods).

Here we validate the assumption of independence between grammatical constraints and cognitive limitations in these classes of grammar using the distance between syntactically related words in a dependency tree as a proxy for memory constraints \citep{Liu2017a,Temperley2018a}. Such a distance is defined as the number of intermediate words plus one. Thus, if the linked words are consecutive they are at distance 1, if they are separated by an intermediate word they are at distance two, and so on, as shown in Fig \ref{syntactic_dependency_trees_figure}. 
Dependency distance minimization is a pressure to reduce the distance between syntactically related words that is supported statistically by large-scale analyses of syntactic dependency structures in languages \citep{Liu2008a,Futrell2015a,Futrell2020a,Jing2021a,Ferrer2020b}. As such, dependency distance minimization is 
a type of memory constraint, believed to result from pressure against decay of activation or interference during the processing of sentences \citep{Liu2017a,Temperley2018a}.
Dependency distances tax memory and cognition in general. Dependency distances reduce in case of cognitive impairment \citep{Roark2011a,Aronsson2021a}. 
There is an association between the level of cognitive impairment and dependency distance: as the severity of the impairment increases, dependency distances tend to be reduced \citep{Aronsson2021a}. 
Moreover, an association between the level of competence of L2 learners and dependency distance has also been found: as learners of a second language become more competent in the new language, dependency distances increase \citep{Ouyang2017a,Yuan2021a}.  

The article is written so that reading the next section, {\em Materials and methods} (Section \ref{materials_and_methods_section}) is not essential to understand the {\em Results} section (Section \ref{results_section}). Therefore, it is up to reader to decide whether to proceed with Section \ref{materials_and_methods_section} or 
to skip to Section \ref{results_section}, reading Section \ref{materials_and_methods_section} later on.


\section{Materials and methods}

\label{materials_and_methods_section}

\subsection*{Control for sentence length}

In our study, we do not investigate the average dependency distance over a whole ensemble of dependency structures but instead we condition on sentence length \citep{Ferrer2013c,Futrell2015a}. Then for a given $n$, we calculate $\left<d\right>_{AS}$, the average dependency length for an ensemble of artificial syntactic dependency structures (AS), and also $\left<d\right>_{RS}$, 
the average dependency length for an ensemble of attested syntactic dependency structures (RS). By doing that, we are controlling for sentence length, getting rid of the possible influence of the distribution of sentence length in the calculation of $\left<d\right>_{RS}$ or $\left<d\right>_{AS}$ \citep{Ferrer2013c}.

\subsection*{Attested syntactic dependency structures}

We estimated the average dependency distances in attested sentences using collections of syntactic dependency treebanks from different languages.  
A syntactic dependency treebank is a database of sentences and their syntactic dependency trees. 

To provide results on a wide range of languages while controlling for the effects of different syntactic annotation theories, we use two collections of treebanks: 
\begin{itemize}
\item Universal Dependencies (UD), version 2.4 \citep{ud24}. This is the largest available collection of syntactic dependency treebanks, featuring 146 treebanks from 83 distinct languages. All of these treebanks are annotated following the common Universal Dependencies annotation criteria, which are a variant of the Stanford Dependencies for English \citep{Stanford2008}, based on lexical-functional grammar \citep{Bresnan00}, adapting them to be able to represent syntactic phenomena in diverse languages under a common framework. This collection of treebanks can be freely downloaded\footnote{\url{https://universaldependencies.org/}. \LastAccessed } and is available under free licenses.
\item HamleDT 2.0 \citep{HamledTStanford}. This collection is smaller than UD, featuring 30 languages, all of which (except for one: Bengali) are also available in UD, often with overlapping source material. Thus, using this collection does not meaningfully extend the diversity of languages covered beyond using only UD. However, the interest of HamleDT 2.0 lies in that each of the 30 treebanks is annotated with not one, but two different sets of annotation criteria: Universal Stanford dependencies \citep{UniversalStanford} and Prague Dependencies \citep{PDT20}. We abbreviate these two subsets of the HamleDT 2.0 collection as ``Stanford'' and ``Prague'', respectively. While Universal Stanford dependencies are closely related to UD, Prague dependencies provide a significantly different view of syntax, as they are based on the functional generative description \citep{Sgall69} of the Praguian linguistic tradition \citep{Hajicova95}, which differs from Stanford dependencies in substantial ways, like the annotation of conjunctions or adpositions \citep{HowFarStanfordPrague}. Thus, using this version of HamleDT\footnote{While there is a later version (HamleDT 3.0), it abandoned the dual annotation and adopted Universal Dependencies instead, thus making it less useful for our purposes.} makes our analysis more robust, as we can draw conclusions without being tied to a single linguistic tradition. The HamleDT 2.0 treebanks are available online.\footnote{\url{https://ufal.mff.cuni.cz/hamledt/hamledt-treebanks-20}. \LastAccessed} While not all of the treebanks are made fully available to the public under free licenses, to reproduce our analysis it is sufficient to use a stripped version where the words have been removed from the sentences for licensing reasons, but the bare trees are available. This version is distributed freely.\footnote{\url{https://lindat.mff.cuni.cz/repository/xmlui/handle/11858/00-097C-0000-0023-9551-4?show=full}. \LastAccessed} 
\end{itemize}
A preprocessed file with the minimal information needed to reproduce our measurements on attested syntactic structures (Fig \ref{lengths_figure}A) is available.
\footnote{
\url{https://doi.org/10.7910/DVN/XHRIYX}
}

To preprocess the treebanks for our analysis, we removed punctuation, following common practice in statistical research of dependency structures \citep{Gomez2016a}. We also removed tree nodes that do not correspond to actual words, such as the null elements in the Bengali, Hindi and Telugu HamleDT corpora and the empty nodes in several UD treebanks. To ensure that the dependency structures are still valid trees after these removals, we reattached nodes whose head has been deleted as dependents of their nearest non-deleted ancestor. Finally, in our analysis we disregarded syntactic trees with less than three nodes, as their statistical properties are trivial and provide no useful information (a single-node dependency tree has no dependencies at all, and a 2-node tree always has a single dependency of distance 1). Table \ref{typological_diversity_table} summarizes the languages in each collection of treebanks. 

\begin{table}
\begin{adjustwidth}{-1.5in}{0in} 
\centering
\caption[Typological diversity of the treebank collections used in this study.]{\label{typological_diversity_table}The languages in every collection grouped by family. The counts attached to the collection names indicate the number of different families and the number of different languages. The counts attached to family names indicate the number of different languages. }
{\scriptsize
\begin{tabularx}{\linewidth}{llX}
Collection & Family & Languages \\
\hline
UD (19, 83) & Afro-Asiatic (7) & Akkadian, Amharic, Arabic, Assyrian, Coptic, Hebrew, Maltese \\
  & Turkik (3) & Kazakh, Turkish, Uyghur \\ 
  & Austro-Asiatic (1) & Vietnamese \\
  & Austronesian (2) & Indonesian, Tagalog \\
  & Basque (1) & Basque \\
  & Dravidian (2) & Tamil, Telugu \\
  & Indo-European (46) & Afrikaans, Ancient Greek, Armenian, Belarusian, Breton, Bulgarian, Catalan, Croatian, Czech, Danish, Dutch, English, Faroese, French, Galician, German, Gothic, Greek, Hindi, Hindi-English, Irish, Italian, Kurmanji, Latin, Latvian, Lithuanian, Marathi, Norwegian, Old Church Slavonic, Old French, Old Russian, Persian, Polish, Portuguese, Romanian, Russian, Sanskrit, Serbian, Slovak, Slovenian, Spanish, Swedish, Ukrainian, Upper Sorbian, Urdu, Welsh \\
  & Japanese (1) & Japanese \\
  & Korean (1) & Korean \\
  & Mande (1) & Bambara \\
  & Mongolic (1) & Buryat \\
  & Niger-Congo (2) & Wolof, Yoruba \\
  & Other (1) & Naija \\
  & Pama-Nyungan (1) & Warlpiri \\
  & Sign Language (1) & Swedish Sign Language \\
  & Sino-Tibetan (3) & Cantonese, Chinese, Classical Chinese \\
  & Tai-Kadai (1) & Thai \\
  & Tupian (1) & Mbya Guarani \\
  & Uralic (7) & Erzya, Estonian, Finnish, Hungarian, Karelian, Komi Zyrian, North Sami \\
Stanford (7, 30) & Afro-Asiatic (1) & Arabic \\
  & Turkik (1) & Turkish \\  
  & Basque (1) & Basque \\
  & Dravidian (2) & Tamil, Telugu \\
  & Indo-European (21) & Ancient Greek, Bengali, Bulgarian, Catalan, Czech, Danish, Dutch, English, German, Greek, Hindi, Italian, Latin, Persian, Portuguese, Romanian, Russian, Slovak, Slovenian, Spanish, Swedish \\
  & Japanese (1) & Japanese \\
  & Uralic (3) & Estonian, Finnish, Hungarian \\
Prague (7, 30) & Afro-Asiatic (1) & Arabic \\
  & Turkik (1) & Turkish \\  
  & Basque (1) & Basque \\
  & Dravidian (2) & Tamil, Telugu \\
  & Indo-European (21) & Ancient Greek, Bengali, Bulgarian, Catalan, Czech, Danish, Dutch, English, German, Greek, Hindi, Italian, Latin, Persian, Portuguese, Romanian, Russian, Slovak, Slovenian, Spanish, Swedish \\
  & Japanese (1) & Japanese \\
  & Uralic (3) & Estonian, Finnish, Hungarian \\

\end{tabularx}
}
\end{adjustwidth}
\end{table}
 
\subsection*{Artificial syntactic dependency structures}

Apart from the attested trees, we used a collection of over 16 billion randomly-generated trees. For values of $n$ (the length or number of nodes) from 3 to $n^*=10$, we exhaustively obtained all possible trees. The number of possible dependency trees for a given length $n$ is given by $n^{n-1}$, ranging from $9$ possible trees for $n=3$ to $10^9$ for $n=n^*$. From $n > n^*$ onwards, the number of trees grows too large to be manageable, so we resort to uniformly random sampling of $10^9$ trees for $n^* < n \leq 25$. 
For each tree in the collection, the classes it belongs to are indicated in the dataset\footnote{
The trees are freely available from 
\url{https://doi.org/10.7910/DVN/XHRIYX}
}. 

\begin{figure}
\centering
\includegraphics[scale = 0.7]{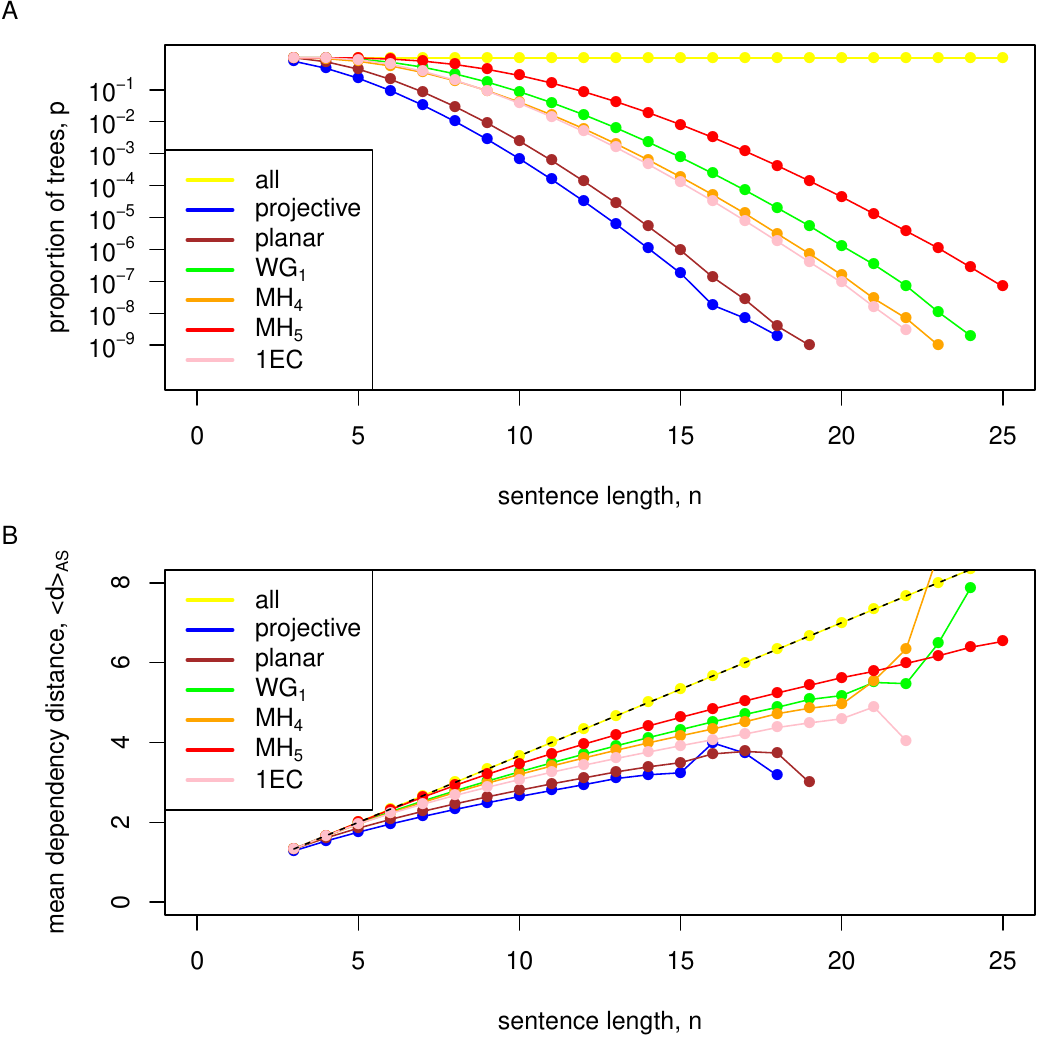}
\caption{\label{undersampling_figure} Undersampling in artificial syntactic dependency structures (AS). A. $p$, the proportion of artificial structures of a certain class in the sample. B. The average dependency length (in words), $\left<d\right>_{AS}$, as a function of $n$, the sentence length (in words). For reference, the base line defined by a random linear arrangement of the words of the sentence, $\left<d\right>_{rla}$ (Eq. \ref{expected_length_equation}) is also shown (dashed line). } 
\end{figure}

The reason why we do not go beyond length 25 is that, for larger lengths, trees that belong to our classes under analysis are very scarce (Fig. \ref{undersampling_figure}A). For example, even sampling $10^9$ random trees for each length, no projective trees are found for $n>18$. The same can be said of planar trees for $n>19$, $1EC$ trees for $n>22$, $MH_4$ trees for $n>23$, and $WG_1$ trees for $n>24$. For the $MH_5$ class, some trees can still be found in the sample for length $25$, but only $69$ out of $10^9$ belong to the class. 
Due to undersampling,  the plot on artificial structures in the results section only shows points represented by at least 30 structures for $n > n^*$. 
30 is considered a rule of thumb for the minimum sample size that is needed to estimate the mean of random variables that follow short tailed distributions \citep{Hogg1997a}. 
Fig. \ref{undersampling_figure}B shows average dependency distances not excluding any point.

For $n \leq n^*$, the ensemble of AS used to calculate $\left<d\right>_{AS}$ contains all possible syntactic dependency structures for all classes. For $n > n^*$, it contains a random sample of them. Within a given ensemble, each structure is generated from a labelled directed tree whose vertex labels are interpreted as vertex positions in the linear arrangement. The values of $\left<d\right>_{AS}$ for each class are exact (the mean over all possible syntactic dependency structures) for $n \leq n^*$ and random sampling estimates for $n > n^*$. A detailed explanation follows.

For a given $n$, an ensemble of syntactic dependency structures is generated with a procedure that is a generalization of the procedure used to generate random structures formed by an undirected tree and a linear arrangement \citep{Esteban2016a}. The procedure has two versions: the exhaustive version, that was used for $n \leq n^*$, and the random sampling version, that was used for $n > n^*$. The exhaustive version consists of
\begin{enumerate}
\item
Generating all the $T(n)$ labelled (undirected) trees of $n$ vertices using Pr\"ufer codes \citep{Pruefer1918a}. It is known that $T(n) = n^{n-2}$ \citep{Cayley1889a}.
\item
Converting each of these random trees into labelled directed trees (i.e., dependency trees) by rooting it in all possible ways. A rooting consists in choosing one node of the tree as the root, and making all edges point away from the root via a depth-first traversal. This produces $nT(n) = n^{n-1}$ syntactic dependency structures.
\item
Producing a syntactic dependency structure from every directed tree using vertex labels (integers from 1 to $n$) as vertex positions in a linear arrangement \citep{Esteban2016a}. 
\item
Discarding the trees that do not belong to the target class.  
\end{enumerate}
The random sampling version consists of 
\begin{enumerate}
\item
Generating $S$ uniformly random labelled (undirected) trees of $n$ vertices, via uniformly random Pr\"ufer codes \citep{Pruefer1918a}.
\item
\label{undirected2directed_step}
Converting these uniformly random labelled trees to uniformly random labelled directed trees (i.e., dependency trees) by randomly choosing one node of each tree as the root, and making all edges point away from the root via a depth-first traversal. This produces $S$ syntactic dependency structures. 
\item
Same as exhaustive version. 
\item
Same as exhaustive version.  
\end{enumerate}
Note that Step \ref{undirected2directed_step} warrants that labelled directed trees in the ensemble are uniformly random: if we call $K_n$ the probability of generating each undirected tree of $n$ vertices with a random Pr\"ufer code, we can observe that each possible directed tree corresponds to exactly one undirected tree (obtained by ignoring arc directions), and each undirected tree corresponds to exactly $n$ distinct directed trees (resulting from picking each of its $n$ nodes as the root). Thus, the method of generating a random Pr\"ufer code and then choosing a root generates each possible directed tree with a uniform probability $K_n/n$ (as the probability of choosing the underlying undirected tree is $K_n$, and the probability of choosing the relevant root is $1/n$).

After each procedure, the average dependency length $\left<d\right>$ for a given $n$ and a given class is calculated. 
While the exhaustive procedure allows one to calculate the true average dependency length over a certain class, the random sampling algorithm only allows one to estimate the true average. 
Put differently, the exhaustive procedure allows one to calculate exactly the expected dependency length in a class assuming that all labelled directed trees are equally likely whereas the random sampling procedure only allows one to obtain an approximation.   
 
We explore all values of $n$ within the interval $[n_{min}, n_{max}]$ with $n_{min} = 3$ and $n_{max} = 25$ and $n^* = 10$ and $S= 10^9$. The total number of syntactic dependency structures generated for our study is
\begin{eqnarray*}
U = (n_{max} - n^*) S + \sum_{n=n_{min}}^{n^*} n T(n) = (n_{max} - n^*) S \sum_{n=n_{min}}^{n^*} n^{n-1}.
\end{eqnarray*}
Applying the parameters above, one obtains
\begin{equation}
U \approx 1.6 \cdot 10^{10}
\end{equation}

\subsection*{The random baseline}

Although the random baseline 
\begin{equation}
\left< d \right>_{rla} = (n+1)/3
\label{expected_length_SM_equation}
\end{equation}
follows from Jaynes' maximum entropy principle in the absence of any constraint \citep{Kesavan2009a}, it may be objected that our baseline is too unconstrained from a linguistic perspective. In previous research, random baselines that assume projectivity or consistent branching, whereby languages tend to grow parse trees either to the right (as in English) or to the left (as in Japanese), have been considered 
\citep{Liu2008a,Gildea2010a,Futrell2015a}. 
However, it has been argued that these linguistic constraints could be a reflection of memory limitations \citep{Ferrer2016a,Christiansen1999a}. Therefore, incorporating these linguistic constraints into the baseline for evaluating dependency distances would not provide an adequate test of the cognitive independence assumption because they could mask the effect of dependency distance minimization (DDm).
Consistently, the planarity assumptions reduces the statistical power of a test of DDm \citep{Ferrer2019a}. In addition, these additional constraints compromise the parsimony of a general theory of language for neglecting the predictive power of DDm \citep{Ferrer2016a}.

{\em A priori}, $\left<d\right>_{AS}$ could be below the random baseline as it occurs typically in human languages \citep{Ferrer2004b,Ferrer2013c} but it could also be above. As for the latter situation, empirical research in short sentences has shown that there are languages where dependency lengths are larger than expected by chance \citep{Ferrer2019a}. In addition, there exist syntactic dependency structures where $\left< d \right> > \left< d \right>_{rla}$ from a network theoretical standpoint. For instance, among planar syntactic structures, the maximum average dependency distance is $\left< d \right>_{max} = n/2$ \citep{Ferrer2013b}.

$\left<d\right>_{AS}$ never exceeds $\left<d\right>_{rla}$ and it deviates from $\left<d\right>_{rla}$ when $n=3$ for projective trees, $n=4$ for planar trees and $MH_4$ and $n=5$ for $1EC$, $MH_5$ and $WG_1$.
For the class of all syntactic dependency structures (Fig. 2 of main article), 
we find that $\left< d \right>_{AS}$ matches Eq. \ref{expected_length_SM_equation} as expected from previous research \citep{Esteban2016a}.

\subsection*{The classes of dependency structures}

{\em Planar trees:} A dependency tree is said to be \emph{planar} (or \emph{noncrossing}) if its dependency arcs do not cross when drawn above the words. 
Planar trees have been used in syntactic parsing algorithms \citep{gomniv2010}, and their generalization to noncrossing graphs has been widely studied both for its formal properties \citep{YliGomACL2017} and for parsing \citep{KuhJon2015}.

\begin{figure}
\centering
\includegraphics[scale = 0.9]{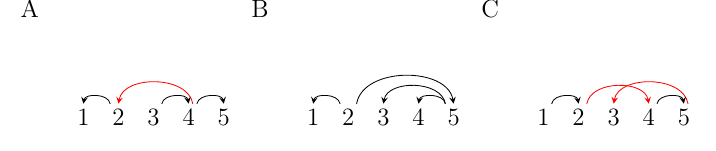}
\caption{\label{prj_pla_figure} Planarity and projectivity. A. A tree that is planar (dependencies do not cross) but not projective (the root node, 3, is covered by the dependency in red). B. A tree that is planar and projective. C. A tree that is not planar (the dependencies in red cross), and thus not projective.} 
\end{figure}

{\em Projective trees:}
A dependency tree is said to be projective if it is planar and its root is not covered by any dependency (see Fig \ref{prj_pla_figure}). Projectivity facilitates the design of simple and efficient parsers \citep{nivre03iwpt,nivre04acl}, whereas extending them to support non-projective trees increases their computational cost \citep{nivre09acl,covington01}. For this reason, and because treebanks of some languages (like English or Japanese) have traditionally had few or no non-projective analyses, many practical implementations of parsers assume projectivity \citep{CheMan2014,Dyer2015}.

However, non-projective parsing is needed to deal with sentences exhibiting non-projective phenomena such as extraposition, scrambling or topicalization. 
Non-projectivity is particularly common in flexible word order languages, but generally present in a wide range of languages. 
However, non-projectivity in natural languages tends to be \emph{mild} in the sense that the actually occurring non-projective trees are very close to projective trees, as they have much fewer crossing dependencies than would be expected by chance \citep{Ferrer2017a}.

For this reason, there has been research interest in finding a restriction that would be a better fit for the phenomena observed in human languages. From a linguistic standpoint, the goal is to describe the syntax of human language better than with the overly restrictive projective trees or the arguably excessive permissiveness of admitting any tree without restriction, disregarding the observed scarcity of crossing dependencies. From an engineering standpoint, the goal is to strike a balance between the efficiency provided by more restrictive parsers with a smaller search space and the coverage of the non-projective phenomena that can be found in attested sentences. In this line, various sets of dependency structures that have been proposed are supersets of projective trees allowing only a limited degree of non-projectivity. These sets are called mildly non-projective classes of dependency trees \citep{KuhNiv06}.

Here, we focus on three of the best known such sets, which have interesting formal properties and/or have been shown to be practical for parsing due to providing a good efficiency-coverage trade-off. We briefly outline them here, and refer the reader to \cite{GomCL2016} for detailed definitions and coverage statistics of these and other mildly non-projective classes of trees.

\begin{figure}
\centering
\includegraphics[scale = 0.9]{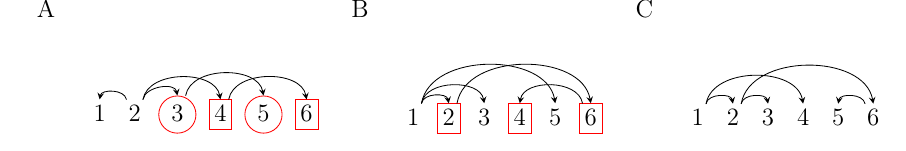}
\caption{\label{wgone_figure} Well-nestedness and gap degree. A. An ill-nested tree (the yields of node 3---circled---and node 4--squared---form an interleaving pattern). B. A tree with gap degree 2 (the yield of node 2, squared, has two discontinuities, at nodes 3 and 5). C. A tree that is well-nested and has gap degree 1, and thus is in $WG_1$.} 
\end{figure}

{\em Well-nested trees with Gap degree 1 ($WG_1$):} A dependency tree is well-nested \citep{bodirsky2005wellnested} if it does not contain two nodes with disjoint, interleaving yields. Given two disjoint yields $a_1 \ldots a_p$ and $b_1 \ldots b_q$, they are said to interleave if there exist $i, j, k, l$ such that $a_i < b_j < a_k < b_l$. On the other hand, the gap degree of a tree is the maximum number of discontinuities present in the yield of a node, i.e., a dependency tree has gap degree 1 if every yield is either a contiguous substring, or the union of two contiguous substrings of the input sentence. Fig \ref{wgone_figure} provides graphical examples of these properties. $WG_1$ trees have drawn interest mainly from the formal standpoint,
for their connections to constituency grammar 
\citep{Kuhlmann2010}, but they also have been investigated in dependency parsing \citep{GomWeiCarEACL,gomez2011cl,corro-wg1}. 

{\em Multi-Headed with at most $k$ heads per item ($MH_k$):} Given $k \ge 3$, the set of $MH_k$ trees contains the trees that can be parsed by an algorithm called $MH_k$ \citep{gomez2011cl}. $k$ is a parameter of the class, such that for $k=3$ the class coincides with projective trees, but for $k>3$ it covers increasingly larger sets of non-projective structures (but the parser becomes slower).
A recent neural implementation of the $MH_4$ parser has obtained competitive accuracy on UD \citep{GomShiLee2018global}. For $k>4$, the $MH_k$ sets have been shown to 
be Pareto optimal (among known mildly non-projective classes) in terms of balance between efficiency and practical coverage
\citep{GomCL2016}. In this paper, we will consider the $MH_4$ and $MH_5$ sets.

\begin{figure}
\centering
\includegraphics[scale = 0.9]{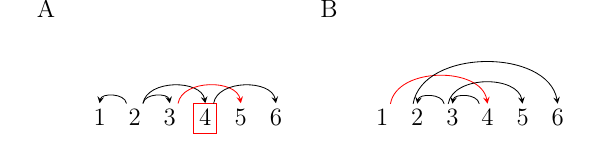}\\
\caption{\label{oneec_figure} 1-Endpoint-Crossing property. A. A 1-Endpoint-Crossing tree (given any dependency, dependencies crossing it are incident to a common node---for example, here the dependencies crossing the one marked in red are incident to node 4). B. A tree that is not 1-Endpoint-Crossing. The dependency arc in red has two crossing dependencies which are not incident to any common node.} 
\end{figure}

{\em 1-Endpoint-Crossing trees ($1EC$):} A dependency tree has the property of being 1-Endpoint-Crossing if, given a dependency, all other dependencies crossing it are incident to a common node \citep{Pitler2013}. This property is illustrated in Fig \ref{oneec_figure}. 1EC trees were the first mildly non-projective class of dependency trees to have a practical exact-inference parser \citep{Pitler2014}, which was reimplemented with a neural architecture in \citep{GomShiLee2018global}. They 
are also in the Pareto frontier with respect to coverage and efficiency, according to \cite{GomCL2016}.

\section{Results}

\label{results_section}

\subsection*{Short dependency distances in attested structures revisited}

Assuming that all the linear arrangements are equally likely, $\left<d \right>$, the average of dependency distances in a sentence of $n$ words, is expected to be \citep{Ferrer2004b}
\begin{equation}
\left< d \right>_{rla} = (n+1)/3.
\label{expected_length_equation}
\end{equation}

\begin{figure}
\centering
\includegraphics[scale = 0.7]{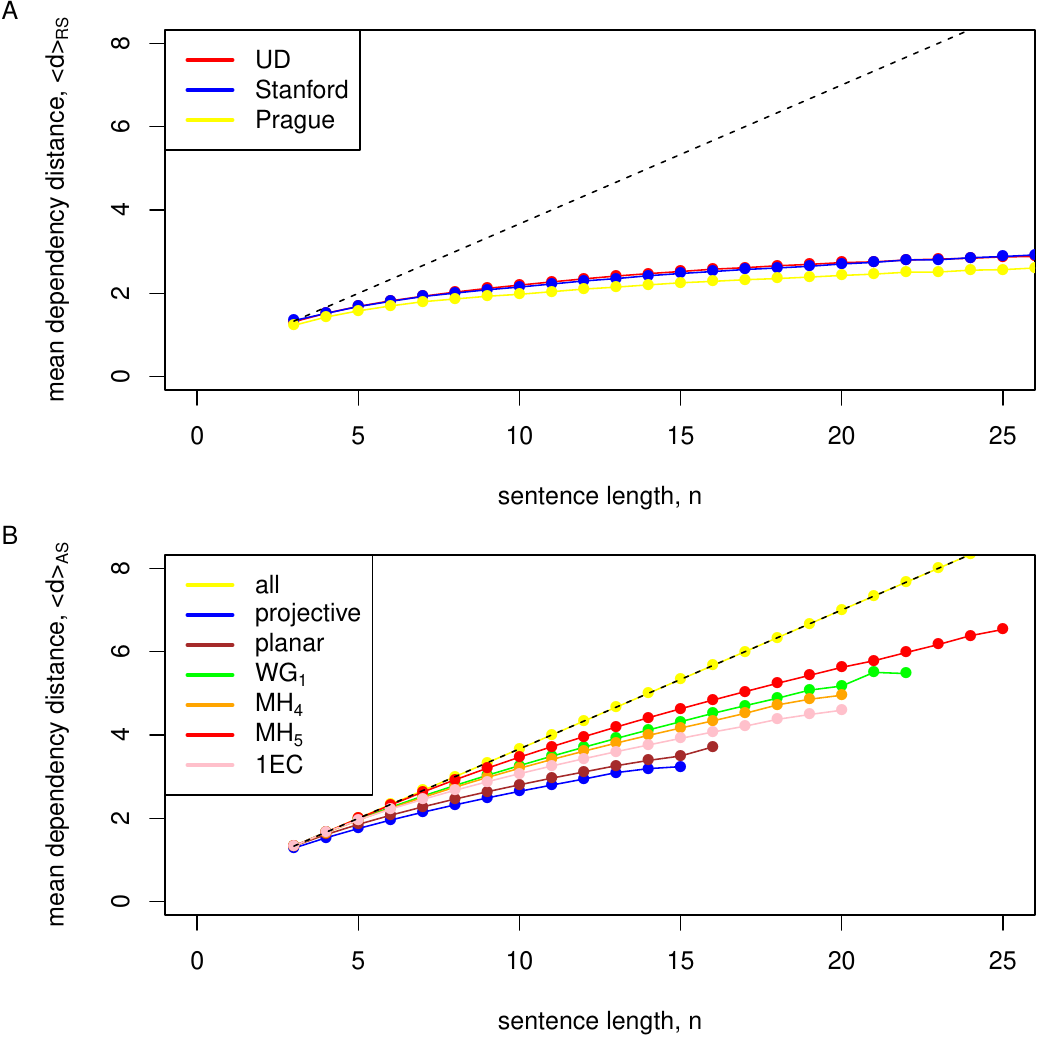}
\caption{\label{lengths_figure} The average dependency length (in words), $\left<d\right>$, as a function of $n$, the sentence length (in words). For reference, the baseline defined by a random linear arrangement of the words of the sentence, $\left<d\right>_{rla}$ (Eq. \ref{expected_length_equation}) is also shown (dashed line). A. Attested syntactic dependency trees (RS) following three different annotation criteria: UD, Prague and Stanford dependencies. B. Artificial syntactic dependency structures (AS) belonging to different classes of grammars. Due to undersampling, only points represented by at least 30 structures are shown for $n > n^*$. }  
\end{figure}

Fig \ref{lengths_figure}A shows that $\left<d\right>_{RS}$, the average dependency distance in attested syntactic dependency structures (RS), is below the random baseline defined by $\left< d \right>_{rla}$ (see Methods for a justification of this baseline). This is in line with 
previous statistical analyses \citep{Ferrer2004b,Liu2008a,Park2009a,Futrell2015a} (see \cite{Liu2017a,Temperley2018a} for a broader review of previous work) and the expected influence of performance constraints on attested trees. 

The fact that $\left<d\right>_{RS}$ is below 4 has been interpreted as a sign that dependency lengths are constrained by working memory limitations \citep{Liu2008a}. 
For this reason, we test whether memory effects have permeated the classes of grammar by determining if $\left<d\right>_{AS}$, the average dependency distance in a collection of artificial syntactic dependency structures (AS) from a certain class, is also below $\left< d \right>_{rla}$ (Eq. \ref{expected_length_equation}). 
The purpose of Fig \ref{lengths_figure}A is merely to provide the reader with a baseline derived from attested dependency structures in natural language as a backdrop for the main contribution of the article, which is based on artificial syntactic dependency structures.

\subsection*{Short dependency distances in artificial structures}

For a given $n$, we generate an ensemble of artificial syntactic dependency structures by exhaustive sampling for $n \leq n^* = 10$ and random sampling for $n > n^*$ (Methods). 
These artificial syntactic dependency trees are only constrained by the definition of the different classes. They are thus free from any memory constraint other than the ones the different classes of grammars may, perhaps, impose indirectly.
Still, these artificial syntactic structures have dependency lengths that are below the chance level (Fig \ref{lengths_figure} B), indicating that memory constraints are hidden in the definition of these classes.
Interestingly $\left<d\right>_{AS}$ 
is below chance for sufficiently large $n$ in all classes of grammars 
although $\left< d \right>_{AS}$ could be above $\left<d\right>_{rla}$ (Eq. \ref{expected_length_equation}) in principle (see Methods). 
In general, the largest reduction of $\left< d \right>_{AS}$ with respect to the random baseline is achieved by the projective class, followed by the planar class.

It is worth noting that a reduction of $\left< d \right>_{AS}$ with respect to our random baseline has been observed for the projective class in past work, but with some important caveats: \cite{Liu2008a} did not control for sentence length as in Fig \ref{lengths_figure} B; and whereas \cite{Park2009a} did implement this control and considered another class of marginal interest (2-component structures) in addition to projective trees, their use of attested dependency trees instead of artificial control trees suggests that memory limitations might have influenced the results.

\section{Discussion}



The reduction of $\left<d\right>$ with respect to the random baseline in artificial trees from a wide range of state-of-the art classes is consistent with the hypothesis that the scarcity of crossing dependencies is a side-effect of pressure to reduce the distance between syntactically related words \citep{Gomez2016a}. 
The smaller reduction of dependency distances with respect to the random baseline in artificial dependency structures can be explained by the fact that the curves in Fig \ref{lengths_figure} B derive from uniform sampling of the space of all possible trees. In contrast, real speakers may not only choose linear arrangements that reduce dependency distance, but also sample the space of possible structures with a bias towards structures that facilitate that such reduction or that satisfy other cognitive constraints \citep{Ferrer2019a}. 

Our findings complete our understanding of the relationship between projectivity or mildly non-projectivity and dependency distance minimization. It has been shown that such minimization leads to a number of edge crossings that is practically zero \citep{Ferrer2006d}, and to not covering the root, one of the conditions for projectivity, in addition to planarity \citep{Ferrer2008e}. Here, we have demonstrated a complementary effect, i.e.,  that dependency distance reduces below chance when edge crossings are minimized (planarity) or projectivity is imposed.
Whereas a recent study of similar classes of grammars suggested that crossing dependencies are constrained by either grammar or cognitive pressures rather than occurring naturally at some rate \citep{Yadav2019a}, our findings strongly demonstrate that it is not grammar but rather non-linguistic cognitive constraints, that limit the 
occurrence of crossing dependencies in languages.
Since we released the first version of this article in August 2019, \url{https://arxiv.org/abs/1908.06629}, other researchers have confirmed that dependency distance minimization contributes significantly to the emergence of formal constraints on crossing dependencies \citep{Yadav2021a, Yadav2022a}. \cite{Yadav2021a} have also confirmed the findings of previous research indicating that the effect of dependency distances alone leads to overestimate the actual number of crossing dependencies \citep{Gomez2016a}; a critical point is that \citet{Gomez2016a} use a normalized score leading to the conclusion that such overestimation implies a small relative error.


We sampled about 16 billion syntactic dependency structures, that differed in length and syntactic complexity, to determine whether linguistic grammars are free of non-linguistic cognitive constraints, as is typically assumed. 
Strikingly, while previous work on natural languages has shown that dependency lengths are considerably below what would be expected by a random baseline without memory constraints  \citep{Ferrer2004b,Liu2008a,Park2009a,Ferrer2013c}, we still observe a drop in dependency lengths for randomly generated, mildly non-projective structures that supposedly abstract away from cognitive limitations.
Our interpretation of these results is that memory constraints, in the form of dependency distance minimization, have become inherent to formal linguistic grammars. 
We have demonstrated that distinct formal classes of mild non-projectivity manifest the sort of burden of dependency distances for memory and cognition that is observed in psychological experiments \citep[Section 2]{Liu2017a} and that has been observed to become more marked in case of cognitive impairment \citep{Aronsson2021a} or second language learning \citep{Ouyang2017a,Yuan2021a}.


It may be objected that our argument that memory limitations have permeated grammars is based on artificially generated syntactic structures instead of real ones. However, 
it is all but impossible to study real dependency structures without possible contamination from linguistic or non-linguistic cognitive constraints other than the formal mild non-projectivity classes. 
For that reason, here and in previous research \citep{Ferrer2014c}, we have focused on artificially generated syntactic structures. Notice this research is part of a larger research program were we have already used real syntactic dependency structures, but minimizing assumptions to argue that the scarcity of crossing dependencies can be explained to a large extent by dependency distance minimization \citep{Gomez2016a}. Nonetheless, further research is needed with real syntactic dependency structures and the current study is a key, necessary step in this direction. 

It may also be objected that our conclusions are limited by the sample of classes that we have considered and that we cannot exclude the possibility that, in the future, researchers might adopt a new class of mildly non-projective structures whose dependency distances cannot be distinguished from the random baseline. However, we believe that this is very unlikely for the following reasons: (1) our current sample of classes is representative of the state of the art \citep{GomCL2016}, and spans classes that originated with different goals and motivations (from purely theoretical to parsing efficiency), with all sharing the drop in dependency lengths, (2) while one could conceivably engineer a class to have lengths in line with the baseline while still having high coverage of linguistic phenomena, this would mean forwarding more responsibility for dependency distance reduction to other parts of the linguistic theory in order to warrant that dependency distances are reduced to a realistic degree (Fig. \ref{lengths_figure}) and hence would preclude a parsimonious approach to language, and (3) given the positive correlation between crossings and dependency lengths \citep{Ferrer2015c,Alemany2019a}, such a class would be likely to have many dependency crossings, so it would be, at the least, questionable to call it mildly non-projective.


Beyond upending longheld assumptions about the nature of human linguistic productivity, our findings also have key implications for debates on how children learn language, how language evolved, and how computers might best master language. Whereas a common assumption in the acquisition literature is that children come to the task of language learning with built-in linguistic constraints on what they learn \citep{Gold1967,Pinker2003a}, our results suggest that language-specific constraints may not be needed and instead be replaced by general cognitive constraints \citep{Tomasello2005}. The strong effects of memory on dependency distance minimization provide further support for the notion that language evolved through processes of cultural evolution shaped by the human brain \citep{Christiansen2008a}, rather than the biological evolution of language-specific constraints \citep{Pinker2003a}. Finally, our results raise the intriguing possibility that if we want to develop computer systems that target human linguistic ability in the context of human-computer interaction, we may paradoxically need to constraint the power of such systems to be in line with human cognitive limitations, rather than giving them the super-human computational capacity of AlphaGo. Memory limitations in the form of dependency minimization have already been applied to machine learning methods, but imposing planarity as if planarity and memory limitations were unrelated constraints \citep[for instance]{Smith2006a,eisner-smith-2010-iwptbook}. This suggests that considering planarity and other formal constraints as the effect of dependency minimization could boost machine learning methods  

Our study was conducted using the framework of dependency grammar, but because of the close relationship between this framework and other ways of characterizing the human unbounded capacity
to produce different sentences \citep{Chomsky1965, Miller1999a}, such as categorial grammar \citep{Morrill2010a}, phrase structure grammar \citep{Gaifman1965,kahane-mazziotta-2015-syntactic}, and minimalist grammar \citep{Osborne2011a}, our results suggest that any parsimonious grammatical framework will incorporate memory constraints. Notice that, as a result of our study, we cannot refute the cognitive independence assumption. Our point is that 
the independence assumption leads to a less parsimonious theory of syntax. We are simply invoking Occam's razor so that formal constraints and the cognitive burden of dependency distances are not treated as separate entities.
Moreover, given that dependency grammars constitute a special case of a graph that is embedded in one dimension, the physics toolbox associated with statistical mechanics and network theory may be applied to provide further insight into the nature of human linguistic productivity \citep{Barthelemy2018a,Gomez2016a}. However, these future explorations notwithstanding, our current findings show that memory limitations have permeated current linguistic conceptions of grammar, suggesting that it may not be possible to adequately capture our unbounded capacity for language, at least in the context of a parsimonious theory compatible with the idea of mild non-projectivity, without incorporating non-linguistic cognitive constraints into the grammar formalism.


\section*{Acknowledgments}


This article is dedicated to the memory of G. Altmann, 1931-2020 \citep{Koehler2021a}.
We are grateful to L. Alemany-Puig, A. Hernandez-Fernandez and M. Vitevitch for helpful commments.
CGR has received funding from the European Research Council (ERC), under the European Union's Horizon 2020 research and innovation
programme (FASTPARSE, grant agreement No 714150), 
from ERDF/MICINN-AEI (ANSWER-ASAP, TIN2017-85160-C2-1-R; SCANNER-UDC, PID2020-113230RB-C21),
from Xunta de Galicia (ED431C 2020/11), and from Centro de Investigaci\'on de Galicia ``CITIC'', funded by Xunta de Galicia and the European Union (ERDF - Galicia 2014-2020 Program), by grant ED431G 2019/01.
RFC is supported by the grant TIN2017-89244-R from MINECO and the recognition 2017SGR-856 (MACDA) from AGAUR (Generalitat de Catalunya).
	
%



\bibliography{../biblio_dt/main,../biblio_dt/twoplanaracl,../biblio_dt/Ramon,../biblio_dt/Ramon_ours,../biblio_dt/twoplanaracl_ours}
\end{document}